# An Alternative Markov Property for Chain Graphs

**Steen A. Andersson**
Department of Mathematics
Indiana University
Bloomington, IN 47405

**David Madigan & Michael D. Perlman**
Department of Statistics, Box 354322
University of Washington
Seattle, WA 98195

## Abstract

Graphical Markov models use graphs, either undirected, directed, or mixed, to represent possible dependences among statistical variables. Applications of undirected graphs (UDGs) include models for spatial dependence and image analysis, while acyclic directed graphs (ADGs), which are especially convenient for statistical analysis, arise in such fields as genetics and psychometrics and as models for expert systems and Bayesian belief networks. Lauritzen, Wermuth, and Frydenberg (LWF) introduced a Markov property for chain graphs, which are mixed graphs that can be used to represent simultaneously both causal and associative dependencies and which include both UDGs and ADGs as special cases. In this paper an alternative Markov property (AMP) for chain graphs is introduced, which in some ways is a more direct extension of the ADG Markov property than is the LWF property for chain graph.

## 1   INTRODUCTION

Graphical Markov models use graphs, either undirected, directed, or mixed, to represent possible dependences among the variables of a multivariate probability distribution. Applications of undirected graphs (UDGs) include models for spatial dependence and image analysis, while acyclic directed graphs (ADGs)[1] occur in genetics, psychometrics, expert systems, Bayesian belief networks, and many other fields. The vertices of the graph represent the variables, while the presence (absence) of an edge between two vertices in-dicates possible dependence (independence) between the two corresponding variables.

[1]The less accurate phrase "directed acyclic graph (DAG)" is more commonly used.

Graphical Markov models determined by ADGs admit especially elegant statistical analysis. The likelihood function associated with an ADG Markov model admits a convenient recursive factorization which, for categorical or multivariate normal data, yields explicit maximum likelihood estimates and likelihood ratio tests - cf. Lauritzen *et al* (1990), Whittaker (1990), Edwards (1995), Lauritzen (1996), Andersson and Perlman (1996). ADG models allow efficient computational algorithms for exact probability calculations, as well as efficient updating algorithms for Bayesian analysis - cf. Pearl (1988), Lauritzen and Spiegelhalter (1988), Shachter and Kenley (1989), Spiegelhalter *et al* (1993).

Lauritzen and Wermuth (1989) and Frydenberg (1990) generalized ADG Markov models to chain graphs; these are graphs with both directed and undirected edges that contain no (partially) directed cycles, and include both UDGs and ADGs as special cases. Chain graph models can be viewed as simultaneously representing dependencies some of which are causal and some associative. Wermuth and Lauritzen (1990), Cox and Wermuth (1993), and Højsgaard and Thiesson (1995) describe statistical applications of chain graphs, while Buntine (1995) discusses their usefulness for modelling belief networks.

It has been noted recently that a chain graph may admit alternative Markov interpretations, hence may simultaneously represent different statistical models $\equiv$ belief networks (Cox and Wermuth (1993, 1996), Wermuth, Cox, and Pearl (1994), Andersson, Madigan, and Perlman (1996b)). These competing Markov interpretations and the assessment of their relative applicability are currently under intensive investigation. In this paper we describe our alternative Markov property (AMP) for chain graphs, which in some ways may be viewed as a more direct generalization of the ADG Markov property than the Lauritzen-Wermuth-Frydenberg (LWF) Markov property for chain graphs.

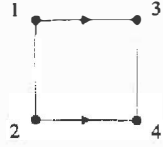

Figure 1: A simple chain graph $G$.

To motivate our AMP, consider the simple chain graph in Figure 1, which represents a set of conditional independences satisfied by random variables $X_1, X_2, X_3, X_4 \equiv 1, 2, 3, 4$. The LWF Markov property (see §3) for $G$ can be expressed in terms of the two conditional independences

$$1 \perp\!\!\!\perp 4 \mid 2, 3, \qquad (1)$$

$$2 \perp\!\!\!\perp 3 \mid 1, 4, \qquad (2)$$

whereas our AMP can be expressed in terms of the two different conditional independences

$$1 \perp\!\!\!\perp 4 \mid 2, \qquad (3)$$

$$2 \perp\!\!\!\perp 3 \mid 1. \qquad (4)$$

Although both interpretations of the chain graph $G$ may be useful for modelling (Cox and Wermuth (1993, p.206)), Cox (1993, p.369) states that "While from one perspective this [LWF property] is easily interpreted, it clearly does not satisfy the requirement of specifying a direct mode of data generation."

By contrast, it is easy to specify a direct mode of data generation for the AMP. Consider, for example the linear simultaneous equations model (SEM)

$$
\begin{aligned}
X_1 &= \epsilon_1 \\
X_2 &= \epsilon_2 \\
X_3 &= b_{31}X_1 + \epsilon_3 \\
X_4 &= b_{42}X_2 + \epsilon_4,
\end{aligned}
$$

where $(\epsilon_1, \epsilon_2)$ and $(\epsilon_3, \epsilon_4)$ are independent random vectors, each with a bivariate normal ($\equiv$Gaussian) distribution with arbitrary correlation, and $b_{31}, b_{42}$ are non-random scalars. Then $(X_1, X_2, X_3, X_4)$ satisfies the AMP conditions (3) and (4) for the chain graph $G$ in Figure 1, but not the LWF conditions (1) and (2) unless $\epsilon_1$ and $\epsilon_2$ are uncorrelated.

This remains true for a general chain graph $G$: our AMP for $G$ is equivalent to the set of conditional independences satisfied by a normal linear block-recursive SEM naturally associated with $G$ (see Remark 4.1), whereas this does not hold in general for the LWF property. (Theorem 4.2 describes those $G$ for which

LWF $\equiv$ AMP.) See Spirtes (1995) and Koster (1996) for related results.

For multivariate normal distributions, the LWF and AMP Markov properties generally are specified by different sets of constraints among regression coefficients and conditional covariance matrices. For example, consider a Gaussian random vector $(X_1, X_2, X_3, X_4)$ with mean vector $(0,0,0,0)$ and positive definite covariance matrix $\Sigma \equiv (\sigma_{ij} \mid i, j = 1, 2, 3, 4)$. It is well known that the conditional distribution of $(X_3, X_4)$ given $(X_1, X_2)$ is the following:

$$
\begin{pmatrix} X_3 \\ X_4 \end{pmatrix} \Big| X_1, X_2 \sim \mathcal{N}_2 \left( \beta \begin{pmatrix} X_1 \\ X_2 \end{pmatrix}, \Lambda \right), \quad (5)
$$

i.e., the bivariate normal distribution with conditional mean

$$
\mathrm{E}\left( \begin{pmatrix} X_3 \\ X_4 \end{pmatrix} \Big| X_1, X_2 \right) = \beta \begin{pmatrix} X_1 \\ X_2 \end{pmatrix} \qquad (6)
$$

and conditional covariance matrix

$$
\mathrm{cov}\left( \begin{pmatrix} X_3 \\ X_4 \end{pmatrix} \Big| X_1, X_2 \right) = \begin{pmatrix} \sigma_{33} & \sigma_{34} \\ \sigma_{43} & \sigma_{44} \end{pmatrix}
$$
$$
- \begin{pmatrix} \sigma_{31} & \sigma_{32} \\ \sigma_{41} & \sigma_{42} \end{pmatrix} \begin{pmatrix} \sigma_{11} & \sigma_{12} \\ \sigma_{21} & \sigma_{22} \end{pmatrix}^{-1} \begin{pmatrix} \sigma_{13} & \sigma_{14} \\ \sigma_{23} & \sigma_{24} \end{pmatrix} =: \Lambda,
$$

where

$$
\beta \equiv \begin{pmatrix} \beta_{31} & \beta_{32} \\ \beta_{41} & \beta_{42} \end{pmatrix} := \begin{pmatrix} \sigma_{31} & \sigma_{32} \\ \sigma_{41} & \sigma_{42} \end{pmatrix} \begin{pmatrix} \sigma_{11} & \sigma_{12} \\ \sigma_{21} & \sigma_{22} \end{pmatrix}^{-1}
$$

is the matrix of regression coefficients of $(X_3, X_4)$ on $(X_1, X_2)$. By (6), the AMP conditions (3) and (4) are equivalent to the directly interpretable conditions

$$\beta_{32} = \beta_{41} = 0,$$

while the LWF conditions (1) and (2) are equivalent to the less directly interpretable conditions

$$\gamma_{32} = \gamma_{41} = 0,$$

where

$$
\begin{pmatrix} \gamma_{31} & \gamma_{32} \\ \gamma_{41} & \gamma_{42} \end{pmatrix} \equiv \gamma := \Lambda^{-1}\beta
$$

is the natural exponential parameter occurring in the conditional normal distribution (5).

Similarly, for a general chain graph model under the assumption of multivariate normality with positive definite covariance matrix, the joint normal distribution factors into a product of conditional normal distributions of the form (5) (not necessarily bivariate), where each conditional distribution involves a regression matrix $\beta$ and a conditional covariance matrix $\Lambda$. Under the AMP, the Markov conditions take the form of separate restrictions on each $\beta$ and on each $\Lambda$ (see Remark 4.1), yielding a *multivariate regression model* (Cox and Wermuth (1993, p.205)). By contrast, under the LWF Markov property, separate restrictions are imposed on each $\gamma := \Lambda^{-1}\beta$ and on each $\Lambda$, yielding a *block regression model.* Again the AMP formulation is perhaps more directly interpretable than the LWF formulation, at least under the assumption of normality. It is somewhat surprising, therefore, that the study of chain graph Markov models has been limited to the LWF interpretation.

The present paper begins a systematic study, continuing in Andersson *et al* (1996b), of the Markov properties of chain graph models under the AMP formulation. For a chain graph that is either a UDG or an ADG, the LWF and AMP Markov properties will be seen to coincide. For a general chain graph, however, the AMP property seems a more direct extension of the ADG Markov property than is the LWF property.

Some graph-theoretic terminology is reviewed in Section 2. A block-recursive Markov property for chain graphs is introduced in Definition 3.3, then, for distributions that satisfy Frydenberg's (1990) condition CI5, shown to be equivalent to the LWF global Markov property in Theorem 3.1. Our alternative block-recursive Markov property for chain graphs is introduced in Definiton 4.1, then shown to be equivalent to a new (AMP) global Markov property in Theorem 4.1, now under *no* restrictions on the distributions. Theorem 4.2 gives the necessary and sufficient condition on a chain graph for its LWF and AMP Markov properties to coincide. In Section 5, some additional properties of our AMP chain graph models are outlined, including the necessary and sufficient condition for their Markov equivalence; these will be discussed in more detail in Andersson *et al* (1996b).

The results in this paper are *not* limited to multivariate distributions that admit joint probability density functions.

# 2    GRAPH-THEORETIC TERMINOLOGY

Our development of graphs and graphical Markov models follows those of Lauritzen *et al* (1990), Frydenberg (1990), and Andersson *et al* (1996a), but with several significant modifications. A *graph* $G$ is a pair $(V, E)$, where $V$ is a finite set of *vertices* and $E \subseteq E^*(V) := \{(v, w) \in V \times V \mid v \neq w\}$ is a set of *edges*, i.e., a set of ordered pairs of distinct vertices. An edge $(v, w) \in E$ whose opposite $(w, v) \in E$ also, is called an *undirected* edge and appears as a line $v—w$ in our figures, whereas an edge $(v, w) \in E$ whose oppo-site $(w, v) \notin E$, is called a *directed* edge and appears as an arrow $v \rightarrow w$. In the text we write $v—w \in G$ and $v \rightarrow w \in G$, respectively. A graph with only undirected edges is called an *undirected graph* (UDG). A graph with only directed edges is a *directed graph* ($\equiv$ *digraph*).

A graph $G' \equiv (V', E')$ is a *subgraph* of $G \equiv (V, E)$, denoted as $G' \subseteq G$, if $V' \subseteq V$ and $E' \subseteq E$. A subset $A \subseteq V$ *induces* the subgraph $G_A := (A, E_A)$, where $E_A := E \cap (A \times A)$; that is, $E_A$ is obtained from $E$ by retaining all edges with both endpoints in $A$. If $B \subseteq A \subseteq V$, clearly $(G_A)_B = G_B$ and $G_B \subseteq G_A$.

Each graph $G = (V, E)$ determines the two UDGs $G^\vee \equiv (V, E^\vee)$, $G^\wedge \equiv (V, E^\wedge)$, defined by

$$E^\vee := \{(v, w) \mid (v, w) \in E \vee (w, v) \in E\},$$
$$E^\wedge := \{(v, w) \mid (v, w) \in E \wedge (w, v) \in E\},$$

respectively. Thus, $G^\vee$ is the *skeleton* of $G$, i.e., the underlying UDG obtained by converting all arrows of $G$ into lines, while $G^\wedge$ is obtained by deleting all arrows of $G$, so $G^\wedge \subseteq G \subseteq G^\vee$. For any subset $A \subseteq V$, $(G_A)^\vee = (G^\vee)_A$ and $(G_A)^\wedge = (G^\wedge)_A$. Two vertices $v, w \in V$ are *adjacent* in $G$ if $(v, w) \in E^\vee$.

The *union* of a finite collection $(G_i \equiv (V_i, E_i) \mid i \in I)$ of graphs is the graph $\cup G_i := (\cup V_i, \cup E_i)$.

Let $G \equiv (V, E)$ be a graph and $A \subseteq V$ a subset of vertices. Denote the *boundary* of $A$ in $G$ by

$$\mathrm{bd}_G(A) := \{v \in V \setminus A \mid (v, a) \in E \text{ for some } a \in A\}$$

and the *closure* of $A$ in $G$ by $\mathrm{cl}_G(A) := \mathrm{bd}(A) \dot{\cup} A$. The *parents* and *neighbors* of $A$ in $G$, denoted by

$$\mathrm{pa}_G(A) := \{v \in V \setminus A \mid v \rightarrow a \in G \text{ for some } a \in A\},$$
$$\mathrm{nb}_G(A) := \{v \in V \setminus A \mid v—a \in G \text{ for some } a \in A\},$$

respectively, are those vertices $b \in V \setminus A$ that are linked to some $a \in A$ in $G$ by directed edges or by undirected edges, respectively. Thus, $\mathrm{bd}_G(A) = \mathrm{pa}_G(A) \cup \mathrm{nb}_G(A)$. The *children* of $A$ in $G$ are defined as

$$\mathrm{ch}_G(A) := \{v \in V \setminus A \mid a \rightarrow v \in G \text{ for some } a \in A\}.$$

We omit $G$ from $\mathrm{bd}_G(A), \mathrm{cl}_G(A), \mathrm{pa}_G(A), \mathrm{nb}_G(A)$, and $\mathrm{ch}_G(A)$, when no confusion could arise.

A *path* of length $n \geq 1$ from $v$ to $w$ in $G$ is an ordered sequence $\{v_0, v_1, \ldots, v_n\}$ of distinct vertices such that $v_0 = v$, $v_n = w$, and $(v_{i-1}, v_i) \in E$ for all $i = 1, \ldots, n$. A *cycle* is a path with the modification that $v_n = v_0$. If $v_{i-1} \rightarrow v_i \in G$ for (all) (at least one) (no) $i$, the path/cycle is called (*directed*), (*semi-directed*), (*undirected*). If $G$ is a UDG (digraph), all paths are undirected (directed).

Let $G \equiv (V, E)$ be a UDG. A subset $A \subseteq V$ is *connected* in $G$ if, for every distinct $a, b \in A$, there is a path between $a$ and $b$ in $G_A$. The maximal connected subsets are called the *connected components* of $G$, and $V$ can be uniquely partitioned into the disjoint union of the connected components of $G$. For pairwise disjoint subsets $A(\neq \emptyset), B(\neq \emptyset)$, and $S$ of $V$, $A$ and $B$ are *separated* by $S$ in the UDG $G$ if all paths in $G$ between $A$ and $B$ intersect $S$. Note that if $S = \emptyset$, then $A$ and $B$ are separated by $S$ in $G$ if and only if there are *no* paths connecting $A$ and $B$ in $G$. In this case, $A$ and $B$ are separated by *any* subset $S$ disjoint from $A$ and $B$.

A graph is called *adicyclic* if it contains no semi-directed cycles. An adicyclic graph is commonly called a *chain graph*. Subgraphs (unions) of chain graphs are (need not be) chain graphs. An *acyclic digraph* (ADG) is a digraph that contains no directed cycles. Thus, UDGs and ADGs are special cases of chain graphs.

For the remainder of this paper, let $G \equiv (V, E)$ be a chain graph. Define the following binary relations on $V$:

$$
\begin{aligned}
v \ll_G w &\iff \exists \text{ a directed path in } G \text{ from } v \text{ to} \\
& \qquad w \in G, \text{ or } v = w; \\
v \leq_G w &\iff \exists \text{ a path in } G \text{ from } v \text{ to } w \in G, \text{ or} \\
& \qquad v = w; \\
v \sim_G w &\iff \exists \text{ an undirected path in } G \text{ from } v \\
& \qquad \text{to } w \in G, \text{ or } v = w \\
&\iff v \leq_G w \text{ and } w \leq_G v.
\end{aligned}
$$

When $G$ is understood, we simply write $v \ll w$, $v \leq w$, and $v \sim w$.

As in Frydenberg (1990), let $\mathcal{T} \equiv \mathcal{T}(G)$ denote the set of equivalence classes in $V$ induced by the equivalence relation $\sim_G$. Equivalently, $\mathcal{T}(G)$ is the set of connected components of $G^\wedge \equiv \cup(G_\tau \mid \tau \in \mathcal{T}(G))$ (see Figures 2a,b). Each vertex $v \in V$ lies in a unique chain component $\tau(v) \in \mathcal{T}$. For each $\tau \in \mathcal{T}$, $\mathrm{nb}(\tau) = \emptyset$, so $\mathrm{bd}(\tau) = \mathrm{pa}(\tau) = \{\mathrm{pa}(v) \mid v \in \tau\}$.

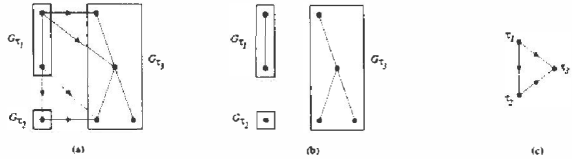

Figure 2: (a) A chain graph $G$ with $\mathcal{T}(G) = \{\tau_1, \tau_2, \tau_3\}$. (b) The UDG $G^\wedge$. (c) The ADG $\mathcal{D}(G)$.

A subset $A \subseteq V$ is called *G-coherent* (or simply *coherent* when $G$ is understood) if $v \in A$ whenever $v \sim a$

for some $a \in A$, that is, if $A$ is a union of chain components of $G$. Equivalently, $A$ is coherent iff $\mathrm{nb}(A) = \emptyset$. If $A$ and $B$ are coherent, then $A \cap B$ is coherent. For any subset $A \subseteq V$, define $\mathrm{Co}(A) \equiv \mathrm{Co}_G(A) :=$ the smallest coherent set containing $A$. Clearly,

$$
\begin{aligned}
\mathrm{Co}(A) &= \{ v \in V \mid v \sim a \text{ for some } a \in A \} \\
&= \cup(\tau \in \mathcal{T} \mid \tau \cap A \neq \emptyset).
\end{aligned}
$$

For any subsets $A, B \subseteq V$, $\mathrm{Co}(A \cup B) = \mathrm{Co}(A) \cup \mathrm{Co}(B)$.

A subset $A \subseteq V$ is called *G-ancestral* (or simply *ancestral* when $G$ is understood) if $v \in A$ whenever $v \ll a$ for some $a \in A$. Equivalently, $A$ is ancestral iff $\mathrm{pa}(A) = \emptyset$. If $A$ and $B$ are ancestral, so is $A \cap B$. Therefore, for any subset $A \subseteq V$, $\mathrm{An}(A) \equiv \mathrm{An}_G(A) :=$ the smallest ancestral set containing $A$, is well-defined and is given by

$$
\mathrm{An}(A) = \{ v \in V \mid v \ll a \text{ for some } a \in A \}.
$$

For any subsets $A, B \subseteq V$, $\mathrm{An}(A \cup B) = \mathrm{An}(A) \cup \mathrm{An}(B)$ (but $\cup$ cannot be replaced by $\cap$).

The *extended subgraph* $G[A]$ determined by a subset $A \subseteq V$ is defined by (see Figures 5 and 6)

$$
G[A] := G_{\mathrm{An}(A)} \cup G^\wedge_{\mathrm{Co}(\mathrm{An}(A))}.
$$

Thus, a *directed* edge occurs in $G[A]$ iff it occurs in $G_{\mathrm{An}(A)}$. If $B \subseteq A \subseteq V$, then $G[B] \subseteq G[A]$.

A subset $A \subseteq V$ is called *G-anterior* (or simply *anterior*) if $v \in A$ whenever $v \leq a$ for some $a \in A$. Equivalently, $A$ is anterior iff $\mathrm{bd}(A) = \emptyset$, so $A$ is anterior iff it is both coherent and ancestral. If $A$ and $B$ are anterior, then $A \cap B$ is anterior. For any subset $A \subseteq V$, define $\mathrm{At}(A) \equiv \mathrm{At}_G(A) :=$ the smallest anterior set containing $A$. Clearly,

$$
\mathrm{At}(A) = \{ v \in V \mid v \leq a \text{ for some } a \in A \}.
$$

For any subsets $A, B \subseteq V$, $\mathrm{At}(A \cup B) = \mathrm{At}(A) \cup \mathrm{At}(B)$ (but $\cup$ cannot be replaced by $\cap$). Note that $\mathrm{An}(A) \subseteq \mathrm{Co}(\mathrm{An}(A)) \subseteq \mathrm{At}(A)$.

For $A \subseteq V$, the subgraph $G(A)$ *spanned by* $A$ is defined by (see Figures 5 and 6)

$$
G(A) := G_{\mathrm{At}(A)};
$$

note that $G[A] \subseteq G(A)$. If $B \subseteq A \subseteq V$, then $G(B) \subseteq G(A)$.

A chain component $\tau \in \mathcal{T}(G)$ is *terminal* in G if $\mathrm{ch}_G(\tau) = \emptyset$. A subset $A \subseteq V$ is *anteriorior* iff it can be generated from $V$ by stepwise removal of terminal chain components. (Note that the removal of a terminal chain component of $G$ might render other chain components terminal in the remaining graph.) If $A$ is anterior and $\tau$ is a terminal chain component of $G$ such that $\tau \subseteq A$, then $\tau$ is a terminal chain component of $G_A$.

The chain components of $G$ themselves comprise the vertices of the graph $\mathcal{D}(G) \equiv (\mathcal{T}(G), \mathcal{E}(G))$, where

$$\mathcal{E}(G) := \{ (\tau, \tau') \in (\mathcal{T}(G) \times \mathcal{T}(G)) \setminus \Delta(\mathcal{T}(G)) \mid \exists\, v \in \tau, v' \in \tau',\ (v, v') \in E \,\}. \quad (7)$$

Then $\mathcal{D} \equiv \mathcal{D}(G)$ is in fact an acyclic digraph (ADG) and $\tau \to \tau' \in \mathcal{D}(G)$ iff $v \to v' \in G$ for some $v \in \tau, v' \in \tau'$ (see Figure 2c). The chain component $\tau$ is terminal in $G$ iff $\tau$ is a terminal vertex in $\mathcal{D}(G)$, i.e., $\mathrm{ch}_{\mathcal{D}(G)}(\tau) = \emptyset$. A subset $A \subseteq V$ is $G$-anterior iff $A = \cup\{\tau \mid \tau \in \mathcal{A}\}$ for some $\mathcal{D}(G)$-ancestral set $\mathcal{A} \subseteq \mathcal{T}(G)$.

If $G \equiv (V, E)$ is a UDG, then $\mathcal{T}(G) = \mathcal{C}(G) :=$ the set of connected components of $G$, $\mathrm{Co}(A) = \mathrm{At}(A)$, $\mathrm{An}(A) = A$, $G[A] = G(A) = G_{\mathrm{Co}(A)}$, $G^{\mathrm{a}} = G^{\mathrm{m}} = G$, and $\mathcal{D}(G) = (\mathcal{C}(G), \emptyset)$.

If $G \equiv (V, E)$ is an ADG, then $\mathcal{T}(G) = V$, $\mathrm{Co}(A) = A$, $\mathrm{An}(A) = \mathrm{At}(A)$, $G[A] = G(A) = G_{\mathrm{An}(A)}$, $G^{\mathrm{a}} = G^{\mathrm{m}}$, and $\mathcal{D}(G) = G$. For $v \in V$, the *descendants* of $v$ in $G$ are defined as follows:

$\mathrm{de}(v) \equiv \mathrm{de}_G(v) :=$
$\{w \in V \setminus \{v\} \mid \exists$ a directed path in $G$ from $v$ to $w\}$.

The *nondescendants* of $v$ are defined by $\mathrm{nd}(v) \equiv \mathrm{nd}_G(v) := V \setminus (\mathrm{de}_G(v) \dot\cup \{v\})$.

An ordered 3-tuple $(a, c, b)$ of distinct vertices of $G$ is called a *flag* in $G$ if the induced subgraph $G_{\{a,c,b\}}$ assumes one of the three forms in Figures 3a,b,c. An ordered 4-tuple $(a, c, d, b)$ of distinct vertices is a *double flag* in $G$ if the induced subgraph $G_{\{a,c,d,b\}}$ has the form in Figure 3d, where the "?" indicates that either $a - b \in G$, $a \to b \in G$, $a \leftarrow b \in G$, or $a$ and $b$ are not adjacent in $G$. In the double flag $(a, c, d, b)$, note that both $(a, c, d)$ and $(c, d, b)$ are flags.

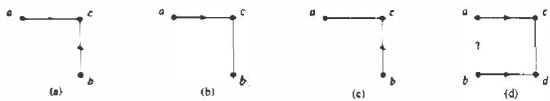

Figure 3: The three configurations that define a flag $(a, c, b)$; the double flag $(a, c, d, b)$.

A flag $(a, c, b)$ is *augmented* by adding the undirected edge $a - b$; a double flag $(a, c, d, b)$ is *augmented* by adding the two undirected edges $a - d$ and $b - c$, and by replacing the "?" by the undirected edge $a - b$. The *augmented graph* $G^{\mathrm{a}}$ derived from a chain graph $G$ is defined to be the UDG obtained by augmenting all flags and double flags in $G$, then converting all directed edges ($\equiv$ arrows) to undirected edges ($\equiv$ lines) - see Figures 4, 5c, and 6c. If $B \subseteq A \subseteq V$, then $G[B]^{\mathrm{a}} \subseteq G[A]^{\mathrm{a}}$.

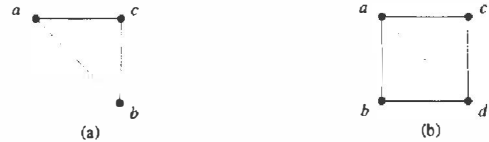

Figure 4: (a) The augmented graph $G^{\mathrm{a}}$ for the chain graphs G in Figures 3a,b,c. (b) The augmented graph $G^{\mathrm{a}}$ for the chain graph G in Figure 3d.

Frydenberg (1990) calls a triple $(a, C, b)$ a *complex* in $G$ if $C$ is a connected subset of a chain component $\tau \in \mathcal{T}(G)$ and $a, b$ are two non-adjacent vertices in $\mathrm{bd}(\tau) \cap \mathrm{bd}(C)$; $(a, C, b)$ is a *minimal complex* in $G$ if no proper subset $C' \subset C$ forms a complex $(a, C', b)$. Every complex $(a, C', b)$ contains at least one minimal complex, $(a, C, b)$ for some $C \subseteq C'$. Frydenberg notes that $(a, C, b)$ is a minimal complex in $G$ if and only if $G_{C \cup \{a,b\}}$ looks like the chain graph of Figure 7. For any subset $A \subseteq V$ such that $C \dot\cup \{a, b\} \subseteq A$, $(a, C, b)$ is a minimal complex in $G_A$ iff it is a minimal complex in $G$. A minimal complex $(a, C, b)$ is a flag iff $C$ contains only one vertex, as in Figure 3a; such a configuration is called an *immorality*.

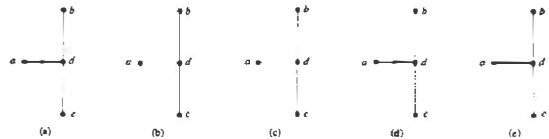

Figure 5: (a) A chain graph G. (b) The extended subgraph $G[\{a, b, c\}]$. (c) The augmented graph $G[\{a, b, c\}]^{\mathrm{a}}$. (d) The subgraph $G(\{a, b, c\})$ spanned by $\{a, b, c\}$. (e) The moral graph $G(\{a, b, c\})^{\mathrm{m}}$.

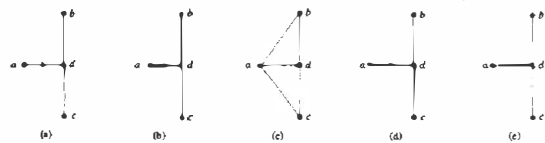

Figure 6: (a) A chain graph G. (b) The extended subgraph $G[\{b, c, d\}]$. (c) The augmented graph $G[\{b, c, d\}]^{\mathrm{a}}$. (d) The subgraph $G(\{b, c, d\})$ spanned by $\{b, c, d\}$. (e) The moral graph $G(\{b, c, d\})^{\mathrm{m}}$.

In Frydenberg's terminology, a minimal complex $(a, C, b)$ is *moralized* by adding the undirected edge $a - b$. The *moral graph* $G^{\mathrm{m}}$ derived from a chain graph $G$ is defined to be the UDG obtained by moralizing all minimal complexes in $G$, then converting all directed edges ($\equiv$ arrows) to undirected edges ($\equiv$ lines) - see Figures 5e, 6e, 8. If $B \subseteq A \subseteq V$, then $G(B)^m \subseteq G(A)^m$. Note that $G[A] \subseteq G(A)$, but neither $G[A]^a \subseteq G(A)^m$ nor $G(A)^m \subseteq G[A]^a$ in general - see Figures 6c,e and 5c,e.

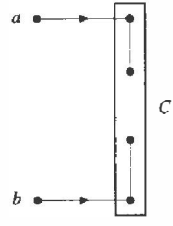

Figure 7: A simple chain graph $G$; $(a, C, b)$ is a minimal complex.

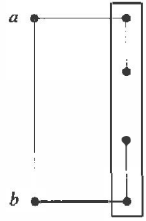

Figure 8: The moral graph $G^m$ for the chain graph $G$ in Fig. 7.

## 3  THE LWF MARKOV PROPERTY FOR CHAIN GRAPHS

We consider multivariate probability distributions $P$ on a product probability space $\mathbf{X} \equiv \times(\mathbf{X}_v | v \in V)$, where $V$ is a finite index set and each $\mathbf{X}_v$ is sufficiently regular to ensure the existence of regular conditional probabilities. Such distributions are conveniently represented by a random variate $X \equiv (X_v | v \in V) \in \mathbf{X}$. For any subset $A \subseteq V$, we define $\mathbf{X}_A := \times(\mathbf{X}_v | v \in A)$, $X_A := (X_v | v \in A)$, and $X_\emptyset :=$ constant. We often abbreviate $X_v$ and $X_A$ by $v$ and $A$, respectively.

For three pairwise disjoint subsets $A, B, C \subseteq V$ and a probability measure $P$ on $\mathbf{X}$, write $A \perp\!\!\!\perp B \mid C [P]$ to indicate that $X_A$ and $X_B$ are conditionally independent given $X_C$ under $P$. Trivially, $A \perp\!\!\!\perp B \mid C [P]$ if $A = \emptyset$ or $B = \emptyset$, while $A \perp\!\!\!\perp B \mid \emptyset [P]$ means that $A$ and $B$ are independent. Dawid (1980) notes that for any four pairwise disjoint subsets $A, B, C, D \subseteq V$,

$$A \perp\!\!\!\perp B \dot\cup C \mid D [P] \iff$$
$$A \perp\!\!\!\perp B \mid C \dot\cup D [P] \text{ and } A \perp\!\!\!\perp C \mid D [P]. \quad (8)$$

A graphical Markov model uses a graph $G \equiv (V, E)$ with vertex set $V$ to specify a Markov property, i.e., a collection of conditional independences, among the component random variates $X_v$, $v \in V$. We shall introduce a certain block-recursive Markov property determined by a chain graph $G$, then establish its equivalence to the LWF global Markov property for distributions that satisfy condition CI5 below. Since this block-recursive Markov property is formulated in terms of the ADG $\mathcal{D} \equiv \mathcal{D}(G) := (\mathcal{T}(G), \mathcal{E}(G)) \equiv (\mathcal{T}, \mathcal{E})$ (cf. (7)) and the family of UDGs $(G_\tau | \tau \in \mathcal{T})$, we first review the local and global Markov properties for ADGs and UDGs, respectively (Lauritzen *et al* (1990)).

**Definition 3.1.** (The local Markov property for ADGs.) Let $D \equiv (V, E)$ be an ADG. A probability measure $P$ on $\mathbf{X}$ is said to be *local D-Markovian* if

$$v \perp\!\!\!\perp (\mathrm{nd}_D(v) \setminus \mathrm{pa}_D(v)) \mid \mathrm{pa}_D(v) [P] \;\; \forall v \in V.$$

**Definition 3.2.** (The global Markov property for UDGs.) Let $G \equiv (V, E)$ be a UDG. A probability measure $P$ on $\mathbf{X}$ is said to be *global G-Markovian* if $A \perp\!\!\!\perp B \mid S [P]$ whenever $S$ separates $A$ and $B$ in $G$.

**Definition 3.3.** (The LWF block-recursive Markov property for chain graphs.) Let $G \equiv (V, E)$ be a chain graph. A probability measure $P$ on $\mathbf{X}$ is said to be *LWF block-recursive G-Markovian* if $P$ satisfies the following three conditions:

(a) $\forall \tau \in \mathcal{T}$, $\tau \perp\!\!\!\perp (\mathrm{nd}_\mathcal{D}(\tau) \setminus \mathrm{pa}_\mathcal{D}(\tau)) \mid \mathrm{pa}_\mathcal{D}(\tau) [P]$, i.e., $P$ is local $\mathcal{D}$-Markovian on $\mathbf{X}$;[2]

(b) $\forall \tau \in \mathcal{T}, \forall \sigma \subseteq \tau$, $\sigma \perp\!\!\!\perp (\mathrm{pa}_\mathcal{D}(\tau) \setminus \mathrm{pa}_G(\sigma)) \mid \mathrm{pa}_G(\sigma) \dot\cup (\tau \setminus \sigma) [P]$;

(c) $\forall \tau \in \mathcal{T}$, the conditional distribution $P_{\tau | \mathrm{pa}_\mathcal{D}(\tau)}$ is global $G_\tau$-Markovian on $\mathbf{X}_\tau$.

The set of all LWF block-recursive $G$-Markovian $P$ on $\mathbf{X}$ is denoted by $\mathcal{P}'_{\mathrm{LWF}}(G; \mathbf{X})$.

**Definition 3.4.** (The LWF global Markov property for chain graphs.) Let $G \equiv (V, E)$ be a chain graph. A probability measure $P$ on $\mathbf{X}$ is said to be *LWF global G-Markovian* if $A \perp\!\!\!\perp B \mid S [P]$ whenever $S$ separates $A$ and $B$ in $G(A \dot\cup B \dot\cup S)^m$. The set of all LWF global $G$-Markovian $P$ on $\mathbf{X}$ is denoted by $\mathcal{P}_{\mathrm{LWF}}(G; \mathbf{X})$.

If $G$ is an ADG, Lauritzen *et al* (1990) showed that its local and LWF global Markov properties are equiva-lent; trivially its LWF block-recursive and LWF global Markov properties coincide in this case. If $G$ is a UDG, Frydenberg (1990, p.339) noted that its global and LWF global Markov properties are equivalent; trivially its LWF block-recursive and LWF global Markov properties also coincide here.

---

[2] Note that $\mathcal{T}$ determines a coarser factorization of the product space $\mathbf{X}$, namely, $\mathbf{X} = \times(\mathbf{X}_\tau | \tau \in \mathcal{T})$.

For *general* chain graphs, Frydenberg (1990, §3) introduced an LWF local Markov property, then showed that the LWF global and LWF local Markov properties are equivalent for probability measures $P$ that satisfy the following condition CI5:[3]

(CI5)    $A \perp\!\!\!\perp B \mid C \dot\cup D\,[P]$ and $A \perp\!\!\!\perp C \mid B \dot\cup D\,[P]$
$$\Rightarrow A \perp\!\!\!\perp (B \dot\cup C) \mid D\,[P]$$

whenever $A, B, C, D$ are pairwise disjoint subsets of $V$. Theorem 3.1(ii) states that under CI5, the LWF global and LWF block-recursive Markov properties are also equivalent. Also see Theorem 2 of Buntine (1995). (All proofs are available in Andersson *et al* (1996b)).

Let $\mathcal{C}$ denote the class of probabilities $P$ on $\mathbf{X}$ that satisfy CI5.

**Theorem 3.1.** (i) $\mathcal{P}_{\text{LWF}}(G; \mathbf{X}) \subseteq \mathcal{P}'_{\text{LWF}}(G; \mathbf{X})$.

(ii) $\mathcal{P}_{\text{LWF}}(G; \mathbf{X}) \cap \mathcal{C} = \mathcal{P}'_{\text{LWF}}(G; \mathbf{X}) \cap \mathcal{C}$.

**Remark 3.1.**    For some $G$'s, $\mathcal{P}_{\text{LWF}}(G; \mathbf{X}) \subset \mathcal{P}'_{\text{LWF}}(G; \mathbf{X})$ - see Andersson *et al* (1996b). By Theorem 4.2 below, such a $G$ must possess at least one flag that is not an immorality.

# 4    AN ALTERNATIVE MARKOV PROPERTY FOR CHAIN GRAPHS

When applied to the chain graph in Figure 1, conditions (a) and (c) are vacuous, while (b) yields the two LWF Markov conditions in (1) and (2). In order to obtain instead the two AMP Markov conditions in (3) and (4), we need only modify condition (b) by deleting the subset $\tau \setminus \sigma$ from the conditioning set, as follows:

(b*)    $\forall \tau \in \mathcal{T}, \forall \sigma \subseteq \tau,\ \sigma \perp\!\!\!\perp (\text{pa}_{\mathcal{D}}(\tau) \setminus \text{pa}_G(\sigma)) \mid \text{pa}_G(\sigma)\,[P]$.

Conditions (a), (b*), and (c) constitute an alternative block-recursive Markov property for chain graphs.

**Definition 4.1.** (The AMP block-recursive Markov property for chain graphs.) Let $G \equiv (V, E)$ be a chain graph. A probability measure $P$ on $\mathbf{X}$ is said to be *AMP block-recursive G-Markovian* if $P$ satisfies conditions (a), (b*), and (c). The set of all AMP block-recursive $G$-Markovian $P$ on $\mathbf{X}$ is denoted by $\mathcal{P}'_{\text{AMP}}(G; \mathbf{X})$.

**Definition 4.2.** (The AMP global Markov property for chain graphs.) Let $G \equiv (V, E)$ be a chain graph. A probability measure $P$ on $\mathbf{X}$ is said to be *AMP global G-Markovian* if $A \perp\!\!\!\perp B \mid S\,[P]$ whenever $S$ separates $A$ and $B$ in $G[A \dot\cup B \dot\cup S]^{\text{a}}$. The set of all AMP global $G$-Markovian $P$ on $\mathbf{X}$ is denoted by $\mathcal{P}_{\text{AMP}}(G; \mathbf{X})$.

**Theorem 4.1.** $\mathcal{P}_{\text{AMP}}(G; \mathbf{X}) = \mathcal{P}'_{\text{AMP}}(G; \mathbf{X})$.

Thus, the AMP global and AMP block-recursive Markov properties are equivalent; here CI5 need *not* be assumed.

The LWF and AMP global Markov properties coincide for UDGs and for ADGs, since

$$G(A \dot\cup B \dot\cup S)^{\text{m}} = G[A \dot\cup B \dot\cup S]^{\text{a}} = G_{\text{Co}(A \dot\cup B \dot\cup S)}$$

if $G$ is a UDG and

$$G(A \dot\cup B \dot\cup S)^{\text{m}} = G[A \dot\cup B \dot\cup S]^{\text{a}} = (G_{\text{An}(A \dot\cup B \dot\cup S)})^{\text{m}}$$

if $G$ is an ADG. The simplest chain graph for which the LWF and AMP global Markov properties differ is the graph $a \rightarrow c\!\!-\!\!b$ consisting of a flag that is not an immorality; here the LWF property is $a \perp\!\!\!\perp b \mid c$, while the AMP property is $a \perp\!\!\!\perp b$. The following result shows that the occurrence of such a flag is necessary and sufficient for the two Markov properties to differ.

**Theorem 4.2.** (i) If $G$ has no flags other than immoralities, then $\forall \mathbf{X}$,

$$\mathcal{P}_{\text{LWF}}(G; \mathbf{X}) = \mathcal{P}_{\text{AMP}}(G; \mathbf{X})$$
$$= \mathcal{P}'_{\text{AMP}}(G; \mathbf{X}) = \mathcal{P}'_{\text{LWF}}(G; \mathbf{X}).$$

(ii) If $G$ has at least one flag that is not an immorality, then for every $\mathbf{X}$ such that $\mathbf{X}_v$ admits a non-degenerate probability measure for each $v \in V$,

$$(\mathcal{P}_{\text{LWF}}(G; \mathbf{X}) \cap \mathcal{C}) \setminus \mathcal{P}_{\text{AMP}}(G; \mathbf{X}) \neq \emptyset \quad (9)$$
$$(\mathcal{P}_{\text{AMP}}(G; \mathbf{X}) \cap \mathcal{C}) \setminus \mathcal{P}_{\text{LWF}}(G; \mathbf{X}) \neq \emptyset. \quad (10)$$

By Theorems 3.1 and 4.1, (9) and (10) remain valid with "$\mathcal{P}$" replaced by "$\mathcal{P}'$".

The chain graph $G$ in Figure 5a has two flags, neither of which is an immorality. By applying (8) repeatedly, it can be shown that $\mathcal{P}_{\text{LWF}}(G; \mathbf{X})$ is determined by the three non-redundant conditions $b \perp\!\!\!\perp c \mid a, d$, $a \perp\!\!\!\perp b \mid d$, and $a \perp\!\!\!\perp c \mid d$, while $\mathcal{P}_{\text{AMP}}(G; \mathbf{X})$ is determined by the three non-redundant conditions $b \perp\!\!\!\perp c \mid a, d$, $a \perp\!\!\!\perp b \mid c$, and $a \perp\!\!\!\perp c$.

**Remark 4.1.** Andersson *et al* (1996b) show that the AMP block-recursive Markov property for a chain graph $G \equiv (V, E)$ is equivalent to the set of conditional independences satisfied by the following recursive normal linear simultaneous equations model:

$$X_\tau = \beta_\tau X_{\text{pa}_{\mathcal{D}}(\tau)} + \epsilon_\tau, \quad \tau \in \mathcal{T}. \tag{11}$$

Here, $X \equiv (X_v | v \in V) \equiv (X_\tau | \tau \in \mathcal{T})$ is a normal random vector in $\mathbf{R}^V$, $\beta_\tau$ is a $\tau \times \text{pa}_{\mathcal{D}}(\tau)$ matrix satisfying

$$v \notin \text{pa}_G(u) \;\Rightarrow\; (\beta_\tau)_{uv} = 0, \tag{12}$$

and $(\epsilon_\tau | \tau \in \mathcal{T})$ is a family of mutually independent normal random vectors with

$$\epsilon_\tau \sim \mathcal{N}(0, \Lambda_\tau), \quad \Lambda_\tau \in \mathbf{P}(G_\tau), \tag{13}$$

where $\mathbf{P}(G_\tau)$ is the set of all positive semidefinite $\tau \times \tau$ covariance matrices such that $\mathcal{N}(0, \Lambda_\tau)$ is global $G_\tau$-Markovian. That $X$ satisfies conditions (a), (b*), and (c) follows from (11), (12), and (13), respectively.

# 5 CONCLUDING REMARKS

An interesting, although complicating, feature of ADG models and chain graph models is the possible non-uniqueness of the graph associated with the model. Unlike undirected graphs, two or more ADGs or chain graphs may determine the same Markov model. For example, the three ADGs $a \to c \to b$, $a \leftarrow c \leftarrow b$, and $a \leftarrow c \to b$ each determine the single Markov condition $a \perp\!\!\!\perp b \mid c$. This non-uniqueness can lead to computational inefficiency in model selection and to inappropriate specification of prior distributions in Bayesian model averaging (Madigan *et al* (1996)).

Two chain graphs $G_1 = (V, E_1)$ and $G_2 = (V, E_2)$ with the same vertex set $V$ are called *LWF (AMP) Markov equivalent* if $\mathcal{P}_{\text{LWF}}(G_1; \mathbf{X}) = \mathcal{P}_{\text{LWF}}(G_2; \mathbf{X})$ $(\mathcal{P}_{\text{AMP}}(G_1; \mathbf{X}) = \mathcal{P}_{\text{AMP}}(G_2; \mathbf{X}))$ for every product space $\mathbf{X}$ indexed by $V$. For ADGs, the two notions of Markov equivalence coincide. Verma and Pearl (1992) prove that two ADGs are Markov equivalent iff they have the same skeleton and same immoralities (also see Madigan (1993)). Frydenberg (1990) and Andersson *et al* (1996a) show that two chain graphs are LWF Markov equivalent iff they have the same skeletons and same minimal complexes. It is shown in Andersson *et al* (1996b) that two chain graphs are AMP Markov equivalent iff they have the same skeletons and same flags. Thus, the condition for AMP Markov equivalence more closely resembles that for ADG Markov equivalence than does the condition for LWF Markov equivalence, in the sense that both immoralities and flags involve only only triples $(a, c, b)$ of vertices, while minimal complexes $(a, C, b)$ can involve arbitrarily many vertices.

The standard computational method used to identify valid conditional independences in ADG models is based on a pathwise separation criterion, called *d-separation*, introduced by Pearl (1988). Bouckaert and Studený (1995) have generalized this to *c-separation*, a more complicated criterion for identifying valid conditional independences in LWF chain graph models. Andersson *et al* (1996b) have obtained a pathwise separation criterion for AMP chain graph models that is simpler than c-separation, due again to the fact that flags involve only triples whereas minimal complexes can be of arbitrary length.

For learning and statistical analysis, chain graphs offer considerable expressive power. Under either the LWF or AMP interpretation, chain graphs can represent many sets of conditional independences that neither ADGs nor UDGs alone can represent. As a consequence, chain graphs encompass many standard statistical model classes (Wermuth and Lauritzen (1990)) and certain neural networks (Buntine (1995)). We speculate that the AMP interpretation will admit simpler Bayesian analysis of chain graph models than will the LWF interpretation, although for both interpretations the formulation of appropriate hyper-Markov laws (Dawid and Lauritzen (1993)) for non-decomposable models remains problematic.

#### Acknowledgements

This research was supported in part by the U. S. National Science Foundation.